\documentclass{article} 
\usepackage{iclr2023_conference,times}

\usepackage{amsmath,amsfonts,bm}

\def\eqref#1{equation~\ref{#1}}

\def\1{\bm{1}}

\DeclareMathAlphabet{\mathsfit}{\encodingdefault}{\sfdefault}{m}{sl}
\SetMathAlphabet{\mathsfit}{bold}{\encodingdefault}{\sfdefault}{bx}{n}

\usepackage{hyperref}
\usepackage{url}

\usepackage{graphicx}
\usepackage{breqn}
\usepackage{xspace}
\usepackage{cleveref}
\crefformat{section}{\S#2#1#3} 
\crefformat{subsection}{\S#2#1#3}
\crefformat{subsubsection}{\S#2#1#3}

\def\name{Chase\xspace}

\title{Chasing Low-Carbon Electricity for Practical and Sustainable DNN Training}

\author{Zhenning Yang, Luoxi Meng, Jae-Won Chung, Mosharaf Chowdhury \\
University of Michigan \\
\texttt{\{znyang,luoxim,jwnchung,mosharaf\}@umich.edu}
}

\iclrfinalcopy 
\begin{document}

\maketitle

\begin{abstract}
Deep learning has experienced significant growth in recent years, resulting in increased energy consumption and carbon emission from the use of GPUs for training deep neural networks (DNNs).
Answering the call for sustainability, conventional solutions have attempted to move training jobs to locations or time frames with lower carbon intensity.
However, moving jobs to other locations may not always be feasible due to large dataset sizes or data regulations. 
Moreover, postponing training can negatively impact application service quality because the DNNs backing the service are not updated in a timely fashion. 
In this work, we present a practical solution that reduces the carbon footprint of DNN training without migrating or postponing jobs.
Specifically, our solution observes real-time carbon intensity shifts during training and controls the energy consumption of GPUs, thereby reducing carbon footprint while maintaining training performance.
Furthermore, in order to proactively adapt to shifting carbon intensity, we propose a lightweight machine learning algorithm that predicts the carbon intensity of the upcoming time frame.
Our solution, \emph{\name}, reduces the total carbon footprint of training ResNet-50 on ImageNet by 13.6\% while only increasing training time by 2.5\%.

\end{abstract}

\section{Introduction}\label{sec:intro}
The growth of Deep Learning has led to a significant increase in energy consumption and carbon emissions from the use of GPUs for training DNNs \citep{Treehouse, SustainableAI, CarbonEmis}, and enhancing the carbon efficiency of DNN training became a pressing and urgent problem.
Concretely, training large DNNs such as GPT-3 \citep{gpt3}, generates 552 metric tons of $\mathrm{CO_2}$ emissions \citep{CarbonEmis}. 

However, not all Joules are born equal; Carbon intensity is a measure of electricity production, and is calculated by considering the number of grams of carbon dioxide emissions produced per kilowatt-hour (kWh) of electricity generated ($\mathrm{g} \cdot \mathrm{CO_{2}}/\mathrm{kWh}$).
Naturally, carbon intensity can vary significantly depending on time and location.
For instance, a region that relies heavily on coal for electricity generation would have a higher carbon intensity \citep{Miller_2022} than one that relies on carbon-free energy sources such as nuclear, solar, or wind \citep{google2018}. 
Additionally, carbon intensity can also vary across time of day or season, as many renewable energy sources depend on natural phenomena.

In this work, we demonstrate that by forecasting and exploiting shifts in real-time carbon intensity, we can enhance the carbon efficiency of DNN training. 
That is, when carbon intensity increases, we slow down training to draw less electricity; on the other hand, when carbon intensity decreases, we speed up training to make more progress.
\emph{\name} makes these decisions automatically and provides large reductions in carbon emissions while increasing training time marginally. 
{\name} will be open-sourced.

\section{Related Work}\label{sec:relatedWork}

Carbon-aware job scheduling utilizes the variation of carbon intensity based on time \citep{jobTime, GeoTimeSol} and location \citep{carbonJobGeo, carbonJobGeo2} in order to reduce the carbon emissions of DNN training. 
But due to various constraints such as large datasets \citep{nuScenes,scannet,ImageNet}, data regulations \citep{gdpr}, and availability of resources, moving jobs to \emph{greener} geographical locations is not always viable. 
Moreover, deferring training jobs to \emph{greener} times may not be an option either, since DNNs must be trained with the latest data and quickly deployed to production for the highest service quality. 
In contrast, our solution does not migrate nor postpone training jobs. Rather, as the training job runs as requested, we transparently adjust its speed and energy consumption so that it automatically \emph{chases} low-carbon electricity.


Optimizing the energy consumption of DNN training can naturally lower carbon emissions due to the linear relationship between carbon and energy. GPUs, the primary hardware used for training DNNs, allow users to set their \emph{power limit} through software \citep{NVML}. 
Exploiting this technique, Zeus~\citep{zeus} jointly optimizes energy and time consumed to reach a target validation accuracy by automatically configuring power limit and batch size over multiple re-training jobs. 
However, Zeus focuses on the time and energy consumption of training jobs and is not aware of carbon intensity nor the time-varying nature thereof.

To proactively react to changes in carbon intensity, having carbon intensity forecasts for the next time window is necessary. 
Recent approaches \citep{recentCarbonForecast2, recentCarbonForecast1} have achieved high forecasting performance, but the use of DNNs consumes GPU resources and can offset the amount of carbon footprint reduction from subsequent optimization techniques. 
On the other hand, there are commercial services \citep{watttime,electricitymaps} that provide historical carbon data and forecasting.
However, the cost of their premium forecasting feature may not be affordable to all. 
We argue that a lightweight and low-cost solution for short-term carbon intensity forecasting is needed to democratize carbon-aware DNN training.




\section{Methodology}\label{sec:method}
In this work, we present a practical approach to reducing the carbon footprint of DNN training. We jointly optimize carbon emission and training performance by tuning the GPU's power limit based on carbon intensity changes, essentially \emph{prioritizing} low-carbon energy.
Moreover, to accurately predict the carbon intensity for the upcoming time window in an affordable manner, we utilize the historical carbon intensity data prior to the training job start time and fit a light regression model. 

\subsection{Carbon Intensity Forecasting}
\label{subsec:forecast}

During training, we aim to periodically adjust the power limit of the GPU by forecasting the carbon intensity until the next invocation.
To build a predictive model for short-term carbon intensity forecasting, when a DNN training job is submitted, historical carbon intensity data one day prior to the start time ($T$ time steps) are retrieved to fit the following regression model
\begin{equation}
 \mathtt{CarbonIntensity}(t) = f(\mathtt{sin\_time}(t),\; \mathtt{cos\_time}(t), \; \mathtt{CarbonIntensity}(t-1))
\end{equation}
where
\begin{equation}
 \mathtt{sin\_time}(t) = \sin \left( \frac{ 2 \pi \cdot t}{T} \right),\;\;\; \mathtt{cos\_time}(t) = \cos \left( \frac{2 \pi \cdot t}{T} \right).
\end{equation}
Our regression model captures the intuition that the carbon intensity of the next time step not only depends on the current carbon intensity but also on the current time of date, which influences the energy mix.
Also, time step $t$ is converted into $\mathtt{sin\_time}(t)$ and $\mathtt{cos\_time}(t)$ to capture the cyclical nature of the diurnal carbon intensity trend.

Users can configure the amount of historical data to collect, and the period between forecasts and power limit adjustments.
For instance, a shorter period will allow more fine-grained power limit tuning, but also invoke forecasting more often.

\subsection{Carbon-Aware DNN Training}

In this section, we develop an online optimization algorithm that adapts the power limit $p$ of the GPU in order to adapt the changing carbon intensity.

The performance of DNN training is often measured by time-to-accuracy (TTA), the time consumed to reach a given target accuracy \citep{tta2019}. We define the carbon emission throughout this process as \textit{carbon-to-accuracy} (CTA):
\begin{dmath}
    \label{eq:CTA}
    \mathtt{CTA} = \mathtt{TTA} \times \mathtt{AvgPower} \times \mathtt{AvgCarbonIntensity}
\end{dmath}
where $\mathtt{AvgPower}$ and $\mathtt{AvgCarbonIntensity}$ are the average power and average carbon intensity during training, respectively. We can formulate our problem as a cost minimization problem over time, where the cost can be defined as:
\begin{dmath}
    \label{eq:Cost}
    \eta \cdot \mathtt{CTA} + (1-\eta) \cdot \mathtt{MaxPower} \cdot \mathtt{MaxCarbonIntensity} \cdot \mathtt{TTA}
\end{dmath}
where $\eta \in [0, 1]$ is a configurable parameter, used to specify the relative importance of carbon efficiency and training performance a priori. GPU $\mathtt{MaxPower}$ and $\mathtt{MaxCarbonIntensity}$ are constants, used to standardize the units of measure in the cost metric ($\mathrm{gCO_2}$) and balance the two terms. 

Through substitution of Equation \ref{eq:CTA} into Equation \ref{eq:Cost}, we obtain the following cost formulation:
\begin{dmath}
    \label{eq:CostSubs}
    \mathtt{TTA} \cdot \left( \eta \cdot \mathtt{AvgPower} \cdot \mathtt{AvgCarbonIntensity} + (1-\eta) \cdot \mathtt{MaxPower} \cdot \mathtt{MaxCarbonIntensity} \right)
\end{dmath}

Solving the full minimization problem directly is difficult due to the difficulty of accurately characterizing two terms in Equation~\ref{eq:CostSubs}:
\begin{enumerate}
  \item $\mathtt{AvgCarbonIntensity}$: While carbon intensity may be predictable for a short time period, it is difficult to reliably predict carbon intensity for the entire duration of training, which could last days to even weeks.
  \item $\mathtt{TTA}$: The stochastic nature of DNN training renders the prediction of TTA very difficult.
\end{enumerate}

Our insight is that carbon intensity will stay relatively constant over a short period of time, providing an opportunity for cost optimization per period. 
Consequently, we propose to iteratively optimize cost in an online manner by forecasting the carbon intensity of each \emph{period} beginning at time step $t \in [1, T]$ (\cref{subsec:forecast}) and determining the optimal GPU power limit $p$ at the beginning of each period.
Thus, for each period, we solve the following optimization problem:
\begin{dmath}
    \label{eq:PeriodCost}
    \min_{p \in \mathcal{P}} \frac{\eta \cdot \mathtt{AvgPower}(p) \cdot \mathtt{CarbonIntensity}(t) + (1-\eta) \cdot \mathtt{MaxPower} \cdot \mathtt{MaxCarbonIntensity}}{\mathtt{Throughput}(p)}
\end{dmath}
where 
$\mathtt{AvgPower}(p)$ represents the profiled power consumption when power limit $p$ is set and $\mathtt{Throughput}(p)$ is inversely proportional to $\mathtt{TTA}$ since changing the power limit of the GPU does not change the number of samples the model will train on.
Our formulation is inspired by Zeus \citep{zeus}, but differs in that we incorporate real-time carbon intensity and adapt to its changes.

To sum up, when a training job arrives, our system first profiles $\mathtt{Throughput}(p)$ and $\mathtt{AvgPower}(p)$ for all $p$ in the set of allowed GPU power limits $\mathcal{P}$. 
Users can specify the length of each period, which determines how often the cost is optimized during the training process. 
At the start of each period, we forecast $\mathtt{CarbonIntensity(t)}$ and determine the optimal power limit for this period by solving Equation~\ref{eq:PeriodCost}. 
Through periodic re-evaluation, we optimize the overall cost of the entire process and make DNN training carbon efficient. 

\section{Results}\label{sec:results}
\subsection{Forecasting Performance}

For evaluation, we have observed that the average change in carbon intensity is less than 0.1\% when retrieved and forecasted in durations shorter than 10 minutes. 
Thus, we retrieved historical carbon intensity trace for the Central US region using the WattTime API~\citep{watttime}, from 2023-01-15 to 2023-01-27 (GMT), with a 30-minute duration (552 data points). 

We found that the first 24 hours of data prior to the DNN training job are sufficient for fitting the regression model. 
The remaining 504 data points or 252 hours were used for testing. 
A list of regression models (Table~\ref{res-table}) was tested. 
To evaluate the performance of the models, we employed the mean absolute percentage error (MAPE) metric commonly used in time series forecasting. 
Support Vector Regression (SVR) \citep{svr} was the best-performing model employed in carbon-aware DNN training. 


\begin{table}[ht]
\caption{Comparison of carbon intensity forecast model performances.}
\label{res-table}
\begin{center}
\begin{tabular}{lr}
Model                     & MAPE \% \\ \hline
\textbf{Support Vector Regression} & \textbf{0.94} \\
Linear Regression         & 1.57 \\
GradientBoosting          & 2.23 \\
AdaBoost                  & 2.51 \\
Random Forest             & 1.76 \\ \hline
\end{tabular}
\end{center}
\end{table}

\subsection{DNN Training}

To evaluate the effectiveness of our solution, we trained ResNet50 \citep{resnet} on the ImageNet dataset \citep{ImageNet} with one NVIDIA A40 GPU. 
$\mathtt{MaxPower}$ is set to 300W, which is the highest possible power limit for the A40 GPU. 
$\mathtt{MaxCarbonIntensity}$ is set to 750$\mathrm{g} \cdot \mathrm{CO_{2}}/\mathrm{kWh}$ which is the observed max intensity within the 24-hour interval prior to the training job start time. 
Our method is compared against Normal Training, which is running the same task with the default GPU configuration (i.e. with $\mathtt{MaxPower}$). 

\begin{figure}[ht]
\begin{center}
\label{res-fig1}
\includegraphics[width=.82\textwidth]{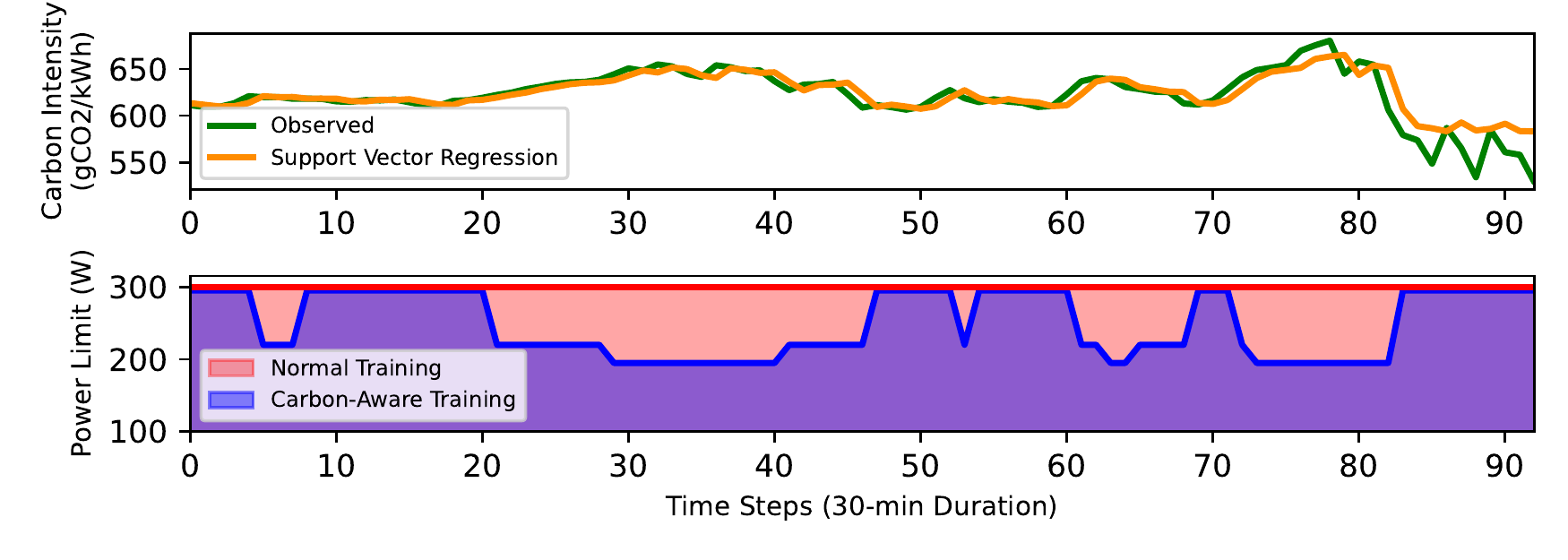}
\end{center}
\caption{The power limit is dynamically adjusted to accommodate for fluctuations in carbon intensity during training. 
    The default power limit for the A40 GPU is 300W. 
    Training with default GPU configuration results in higher energy consumption and subsequently higher carbon emissions.}
\end{figure}

\begin{figure}[ht]
\begin{center}
\label{res-fig2}
\includegraphics[width=.82\textwidth]{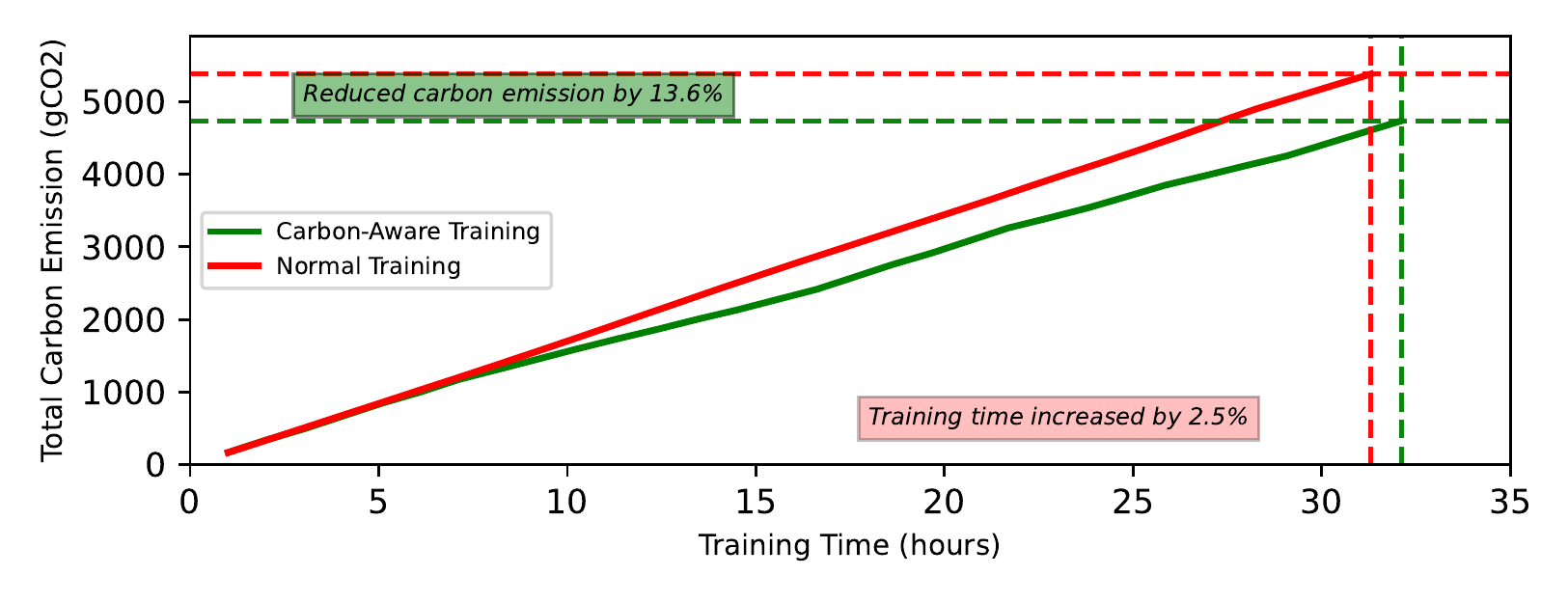}
\end{center}
\caption{In comparison to Normal Training, Carbon-Aware Training reduces carbon emissions during the entire training process and achieves the same accuracy with marginally longer training time.}
\end{figure}

Our solution effectively reduces the total carbon footprint by 13.6\% compared to normal DNN training methods (Figure \ref{res-fig2}). 
This is achieved through the use of less electricity and dynamic power limit adjustments to prioritize greener energy sources (Figure \ref{res-fig1}), with only a minimal increase of 2.5\% in training time, 
allowing even time-sensitive DNN training jobs to reduce carbon emissions immediately.

\section{Conclusion}\label{sec:conclusion}
In conclusion, this work addresses the problem of reducing the energy consumption and carbon emissions of DNN training on GPUs.
By utilizing a simple regression model and a limited amount of historical data, we demonstrate that high short-term forecasting performance can be achieved. 
By incorporating this information, our solution dynamically and automatically adjusts the GPU power limit in real time, reducing carbon emissions without the need for job migration or deferral. 
As future work, we believe that extending {\name} to support multiple DNN training jobs in data centers can significantly contribute to the fight against climate change. 


\section{Acknowledgements}\label{sec:Acknowledgements}
We would like to thank the reviewers and SymbioticLab members for their insightful feedback. 
This work is in part supported by NSF grants CNS-1909067 and CNS-2104243 and a grant from VMWare.
Jae-Won Chung is additionally supported by the Kwanjeong Educational Foundation.

\bibliography{iclr2023_conference}
\bibliographystyle{iclr2023_conference}


\end{document}